\pdfoutput=1

\documentclass[11pt]{article}

\usepackage{EACL2023}

\usepackage{times}
\usepackage{latexsym}

\usepackage[T1]{fontenc}

\usepackage[utf8]{inputenc}

\usepackage{microtype}

\usepackage{inconsolata}
\usepackage{graphicx}

\usepackage[colorinlistoftodos,prependcaption,textsize=small]{todonotes}

\usepackage{caption}
\usepackage{subcaption}
\usepackage{float}

\title{NxPlain: A Web-based Tool for Discovery of Latent Concepts}

\author{
 Fahim Dalvi ~ Nadir Durrani ~ Hassan Sajjad$^{\clubsuit}$\thanks{\hspace{1.5mm} This work was carried out while the author was at QCRI.} \\  \textbf{Tamim Jaban} ~ \textbf{Mus'ab Husaini} ~ \textbf{Ummar Abbas} \\ 
{\tt \{faimaduddin,ndurrani\}@hbku.edu.qa} \\ 
Qatar Computing Research Institute, HBKU Research Complex, Doha, Qatar \\
\textsuperscript{$\clubsuit$}Faculty of Computer Science, Dalhousie University, Halifax, Canada  \\ 
}

\begin{document}
\maketitle
\begin{abstract}

The proliferation of deep neural networks in various domains has seen an increased need for the interpretability of these models, especially in scenarios where fairness and trust are as important as model performance. A lot of independent work is being carried out to: i) analyze what linguistic and non-linguistic knowledge is learned within these models, and ii) highlight the salient parts of the input. We present \textbf{NxPlain}, a web application that provides an explanation of a model's prediction using latent concepts. NxPlain discovers latent concepts learned in a deep NLP model, provides an interpretation of the knowledge learned in the model, and explains its predictions based on the used concepts. The application allows users to browse through the latent concepts in an intuitive order, letting them efficiently scan through the most salient concepts with a global corpus-level view and a local sentence-level view. Our tool is useful for debugging, unraveling model bias, and for highlighting spurious correlations in a model. A hosted demo is available here: \url{https://nxplain.qcri.org}\footnote{A short video demo of the system is also available here: \url{https://www.youtube.com/watch?v=C2PiO4fI5dk}}


\end{abstract}

\section{Introduction}

Interpretation of deep neural networks (DNNs) has gained a lot of attention in recent years, especially in NLP, where state-of-the-art models are being widely deployed and used in practice. Work done in interpretation can be broadly classified into two branches: i) representation analysis and ii) attribution analysis. The former attempts to understand what knowledge is learned within the representation ~\cite{belinkov:2017:acl, tenney-etal-2019-bert} and the latter is focused on how the model predicts the output ~\cite{linzen_tacl,gulordava-etal-2018-colorless,marvin-linzen-2018-targeted}.\footnote{The following survey papers summarize the work done on \emph{Representations Analysis} \cite{belinkov-etal-2020-analysis,neuronSurvey} and \emph{Attribution Analysis} \cite{danilevsky-etal-2020-survey}}

\begin{figure*}
     \centering
     \begin{subfigure}[t]{0.3\textwidth}
         \centering
         \includegraphics[width=\textwidth]{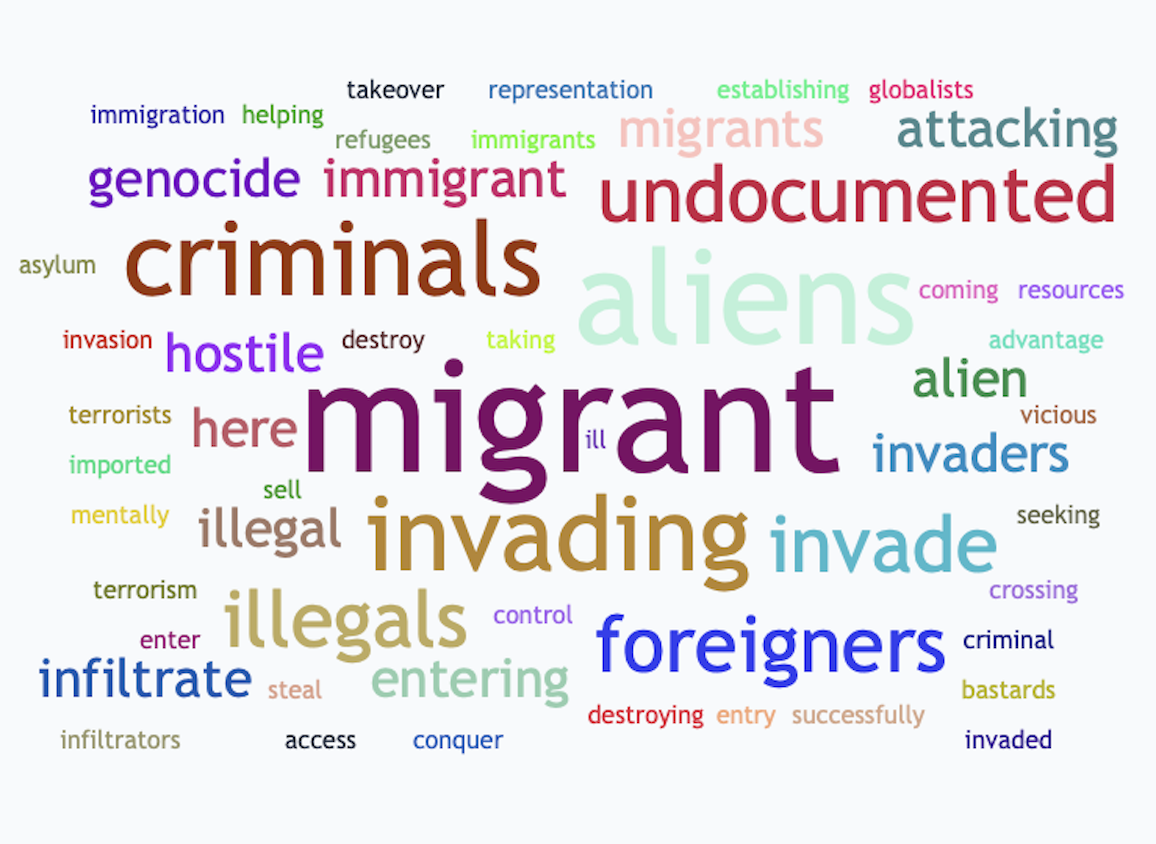}
         \caption{Terms used in hate-speech against immigration policies}
         \label{fig:concept-alien}
     \end{subfigure}
     \hfill
     \begin{subfigure}[t]{0.3\textwidth}
         \centering
         \includegraphics[width=\textwidth]{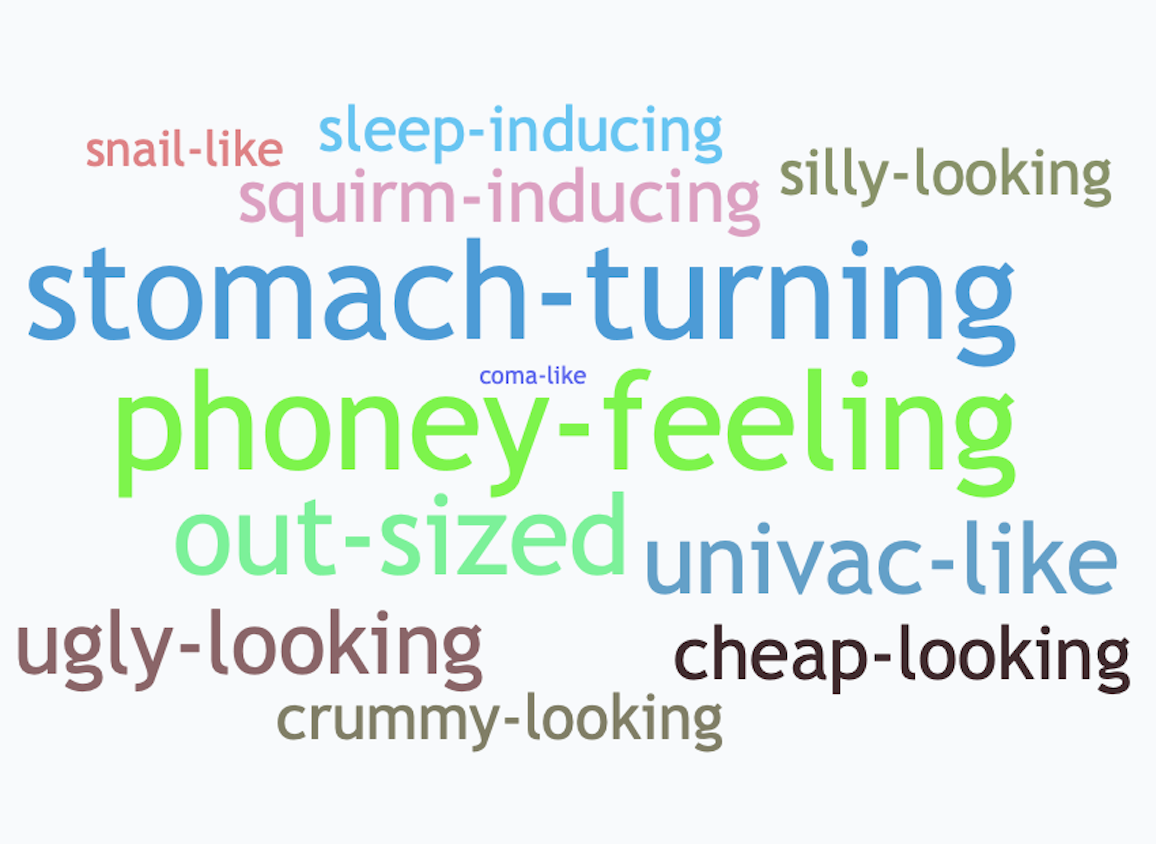}
         \caption{Syntactic concept of hyphenated words}
         \label{fig:concept=hyphenated}
     \end{subfigure}
     \hfill
     \begin{subfigure}[t]{0.3\textwidth}
         \centering
         \includegraphics[width=\textwidth]{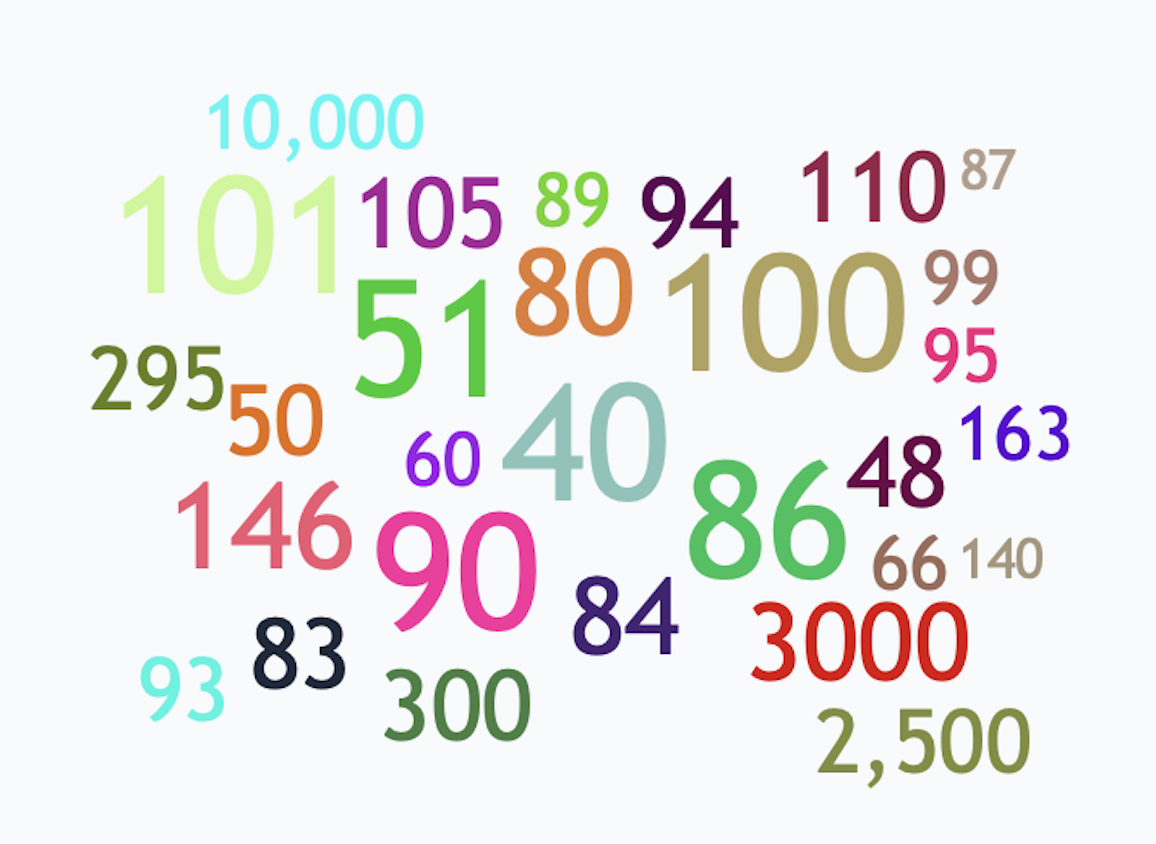}
         \caption{Concept made up of numbers}
         \label{fig:concept-numbers}
     \end{subfigure}
     \begin{subfigure}[t]{0.3\textwidth}
         \centering
         \includegraphics[width=\textwidth]{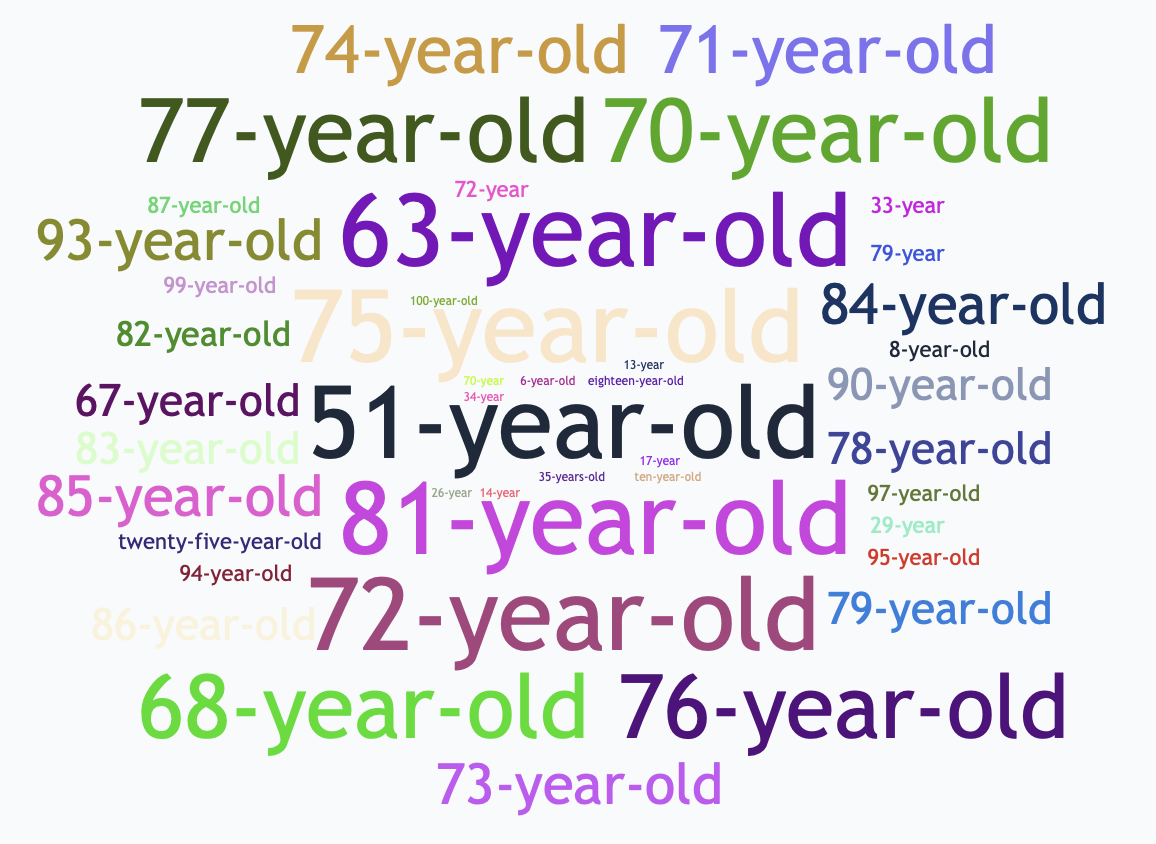}
         \caption{Lexical concept (hyphenation) representing ages}
         \label{fig:concept-ages}
     \end{subfigure}
     \hfill
     \begin{subfigure}[t]{0.3\textwidth}
         \centering
         \includegraphics[width=\textwidth]{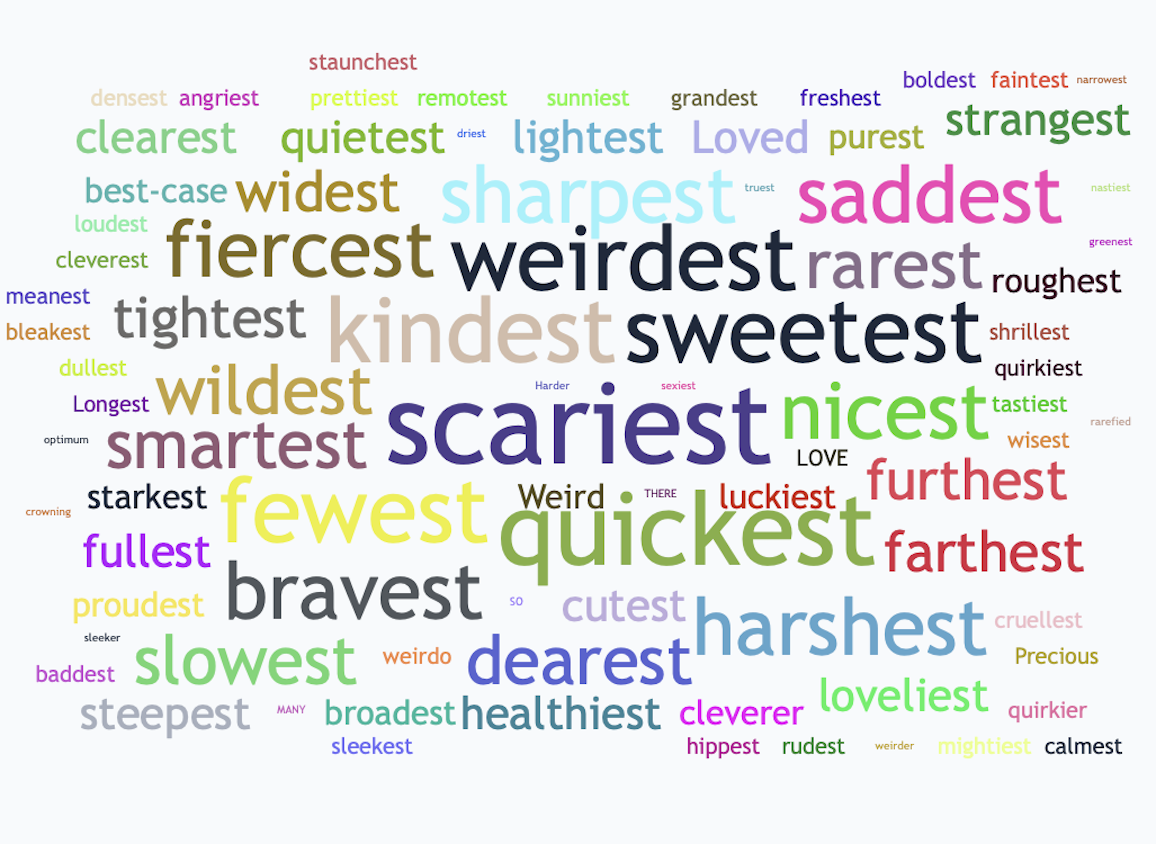}
         \caption{Morphological concept (\textit{adjectives} with common suffix \textit{est} signifying superlative adjectives}
         \label{fig:concept=adjectives}
     \end{subfigure}
     \hfill
     \begin{subfigure}[t]{0.3\textwidth}
         \centering
         \includegraphics[width=\textwidth]{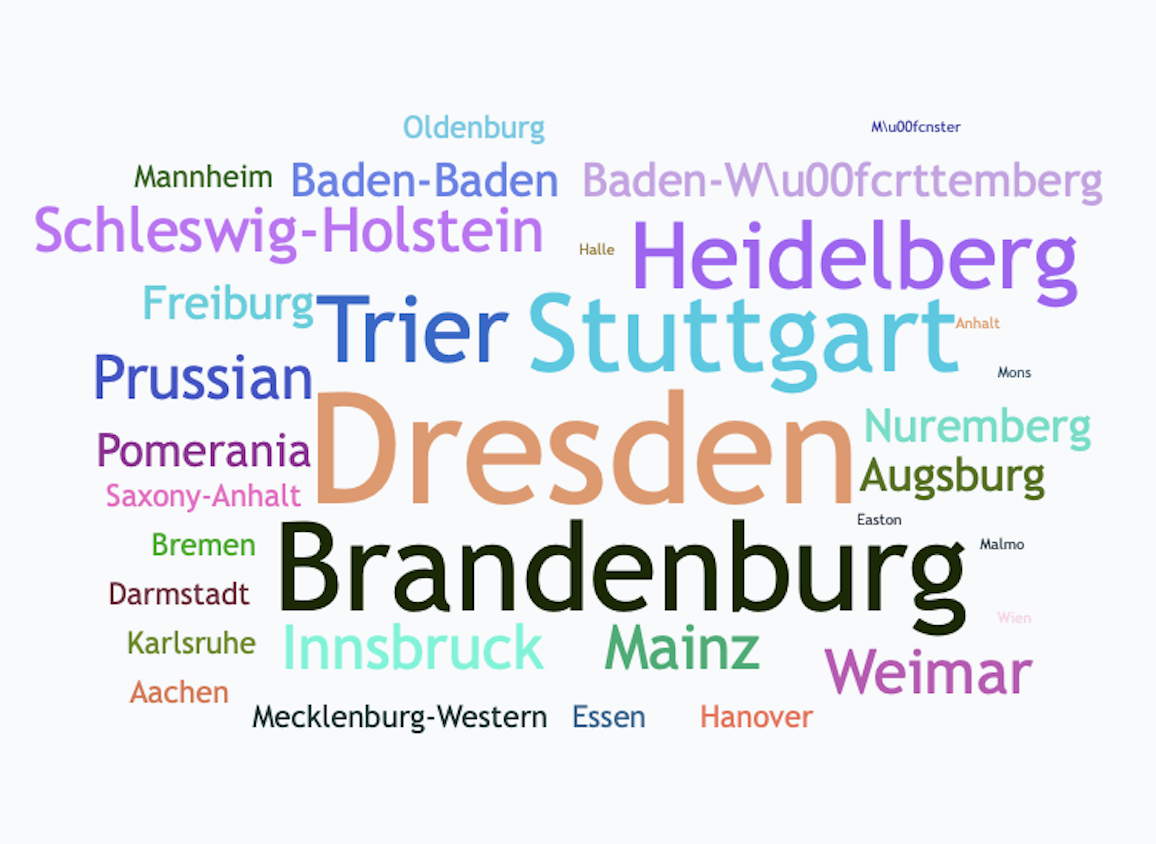}
         \caption{Named entities in Germany}
         \label{fig:concept-germany}
     \end{subfigure}
    \caption{Examples of Latent Concepts.}
    \label{fig:concept-examples}
\end{figure*}

A drawback of the methods in \textit{representation analysis} is that it does not gauge whether the model uses what it has learned in making a prediction. On the other hand, the drawback of \textit{attribution analysis} is that their explanations are limited to discrete units (e.g. words, some specific piece of the network), and the abstract nuances behind these discrete units are lost in the explanation, resulting in an inadequate or implausible explanation. Some efforts have been made in trying to connect representation and attribution analysis~\cite{feder-etal-2021-causalm, elazar-etal-2021-amnesic}. 

In this work, we present \textbf{NxPlain}, a web-app that provides a holistic view by combining representation and attribution analysis. More specifically, we discover latent concepts in the model using the Latent Concept Analysis~\cite{dalvi2022discovering} and connect these concepts to specific predictions using Integrated Gradients~\cite{SundararajanTY17}, a model and input saliency method. \\

\textbf{NxPlain} allows the users to:

\begin{itemize}
    \item Discover latent concepts in \emph{transformers}~\cite{wolf-etal-2020-transformers} models via an interactive GUI
    \item Align the concepts using human-defined ontologies and task specific concepts
    \item Explain predictions using saliency-based attributions and extracted latent concepts
\end{itemize}

The analysis presented by \textbf{NxPlain} can enable a practitioner to understand a trained model better and be aware of the kinds of concepts a model is using to perform its tasks. For example, the word \emph{immigrant} can appear as part of a neutral concept (if the model clusters it with other "roles" related to a person's status like "non-immigrant", "resident", etc), or it can appear as part of a negative concept (if the model clusters it with other hate-speech related terms like "alien", "illegal" etc.) as in Figure \ref{fig:concept-examples}. Understanding which of these categorizations a model is learning and relying on can be a strong signal of the underlying biases of the model. A more benign example of debugging would also be able to see a purely lexical concept being used for prediction (say words ending in "y"), when the lexical property should not have any bearing on the task at hand. The target users for our system can be broadly divided into two categories: i) researchers/practitioners who want to understand their model better, and ii) other systems that want to use the concepts extracted by NxPlain to better explain predictions to their customers.

\begin{figure*}[!ht]
    \centering
    \includegraphics{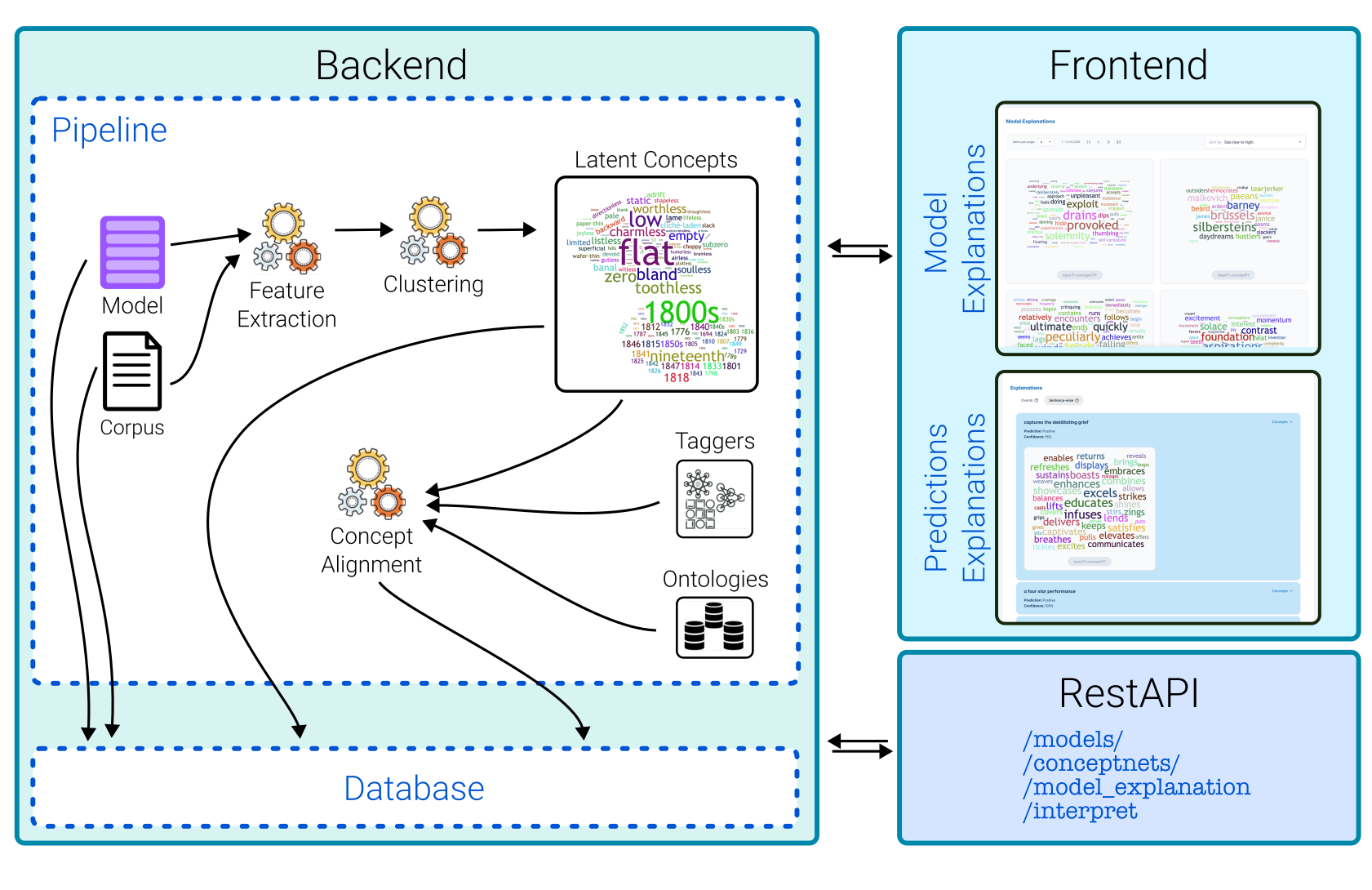}
    \caption{The architecture of NxPlain: The backend uses a pipeline to extract latent concepts and align them with various human ontologies and task-specific concepts. The frontend then uses the computed data to provide both global (model-level) and local (prediction-level) explanations. A RestAPI is also provided so a user can build upon the backend without having to use the provided frontend.}
    \label{fig:architecture}
\end{figure*}

\section{System Design}

The overall system behind the \textbf{NxPlain} application is split into three distinct components. See Figure \ref{fig:architecture} for a pictorial representation.
\begin{itemize}
    \item \textbf{\texttt{Backend}}: This part of the app integrates the \texttt{pipeline}, which handles i) extraction of latent concepts, ii) computation of various orderings, and iii) computation of the concepts relevant to particular sentences etc. A database is used to store all of the computed results so that the other two components can then use these results.
    \item \textbf{\texttt{Rest API}}: This piece displays the results from the \texttt{Backend} in an organized and machine-readable fashion. Users can use this to access the latent concepts and their relevant metadata for their applications.
    \item \textbf{\texttt{Frontend}}: This is the primary user-facing module of the app, and runs in a Web browser. The \texttt{frontend} provides an easy to use the graphical interface to add models to the computation queue and retrieve the extracted concepts once they are ready. Figure \ref{fig:model-explanations} shows the \emph{Model Explanations} page, where one can browse all the extracted concepts, sort them according to various criteria and analyze the knowledge learned in the selected model.
\end{itemize}

\paragraph{Technical Details} For extracting the concepts, we use the code provided by \citet{dalvi2022discovering}. We then tag the input corpus with various human-defined tagsets such as Parts-of-Speech and Semantic tags, and align the latent concepts with these, as done by \citet{sajjad-etal-2022-analyzing}. The results are then stored in a database, and retrieved later via a Python server implemented using Flask. The backend exposes a Rest API which can be used as-is by users in their own applications. We also provide an Angular frontend app that uses the Rest API to present the concepts in a GUI. For sentence-level explanations, we use the \cite{Captum} tookit's Integrated Gradients implementation to perform attribution analysis.

\section{Pipeline Components}
\label{sec:research-modules}

The \textbf{NxPlain} application provides an easy interface to analyze the latent knowledge learned within a deep NLP model, as well as connect these latent concepts to specific predictions. In order to do this, the pipeline in the \texttt{Backend} relies on three key components proposed by recent literature: i) concept discovery, ii) concept alignment, and iii) attribution analysis. 

\subsection{Concept Discovery}
The first component, responsible for extracting the latent concepts learned by a model is based on work done by \newcite{dalvi2022discovering}, called \textit{Latent Concept Analysis}. At a high level, feature vectors (contextualized representations) are first generated by performing a forward pass on the model. These representations are then clustered using agglomerative hierarchical clustering~\citep{gowda1978agglomerative} to discover the encoded concepts. The hypothesis is that contextualized word representations learned within pretrained language models capture \emph{meaningful} groupings based on a coherent concept such as lexical, syntactic and semantic similarity, or any task or data specific pattern that groups the words together \cite{dalvi2022discovering}. Figure \ref{fig:concept-examples} shows example concepts discovered in the model space of a base and finetuned BERT model. The concepts discovered are a mix of linguistic, lexical and semantic concepts.

\begin{figure*}[!ht]
    \centering
    \includegraphics[width=0.99\linewidth]{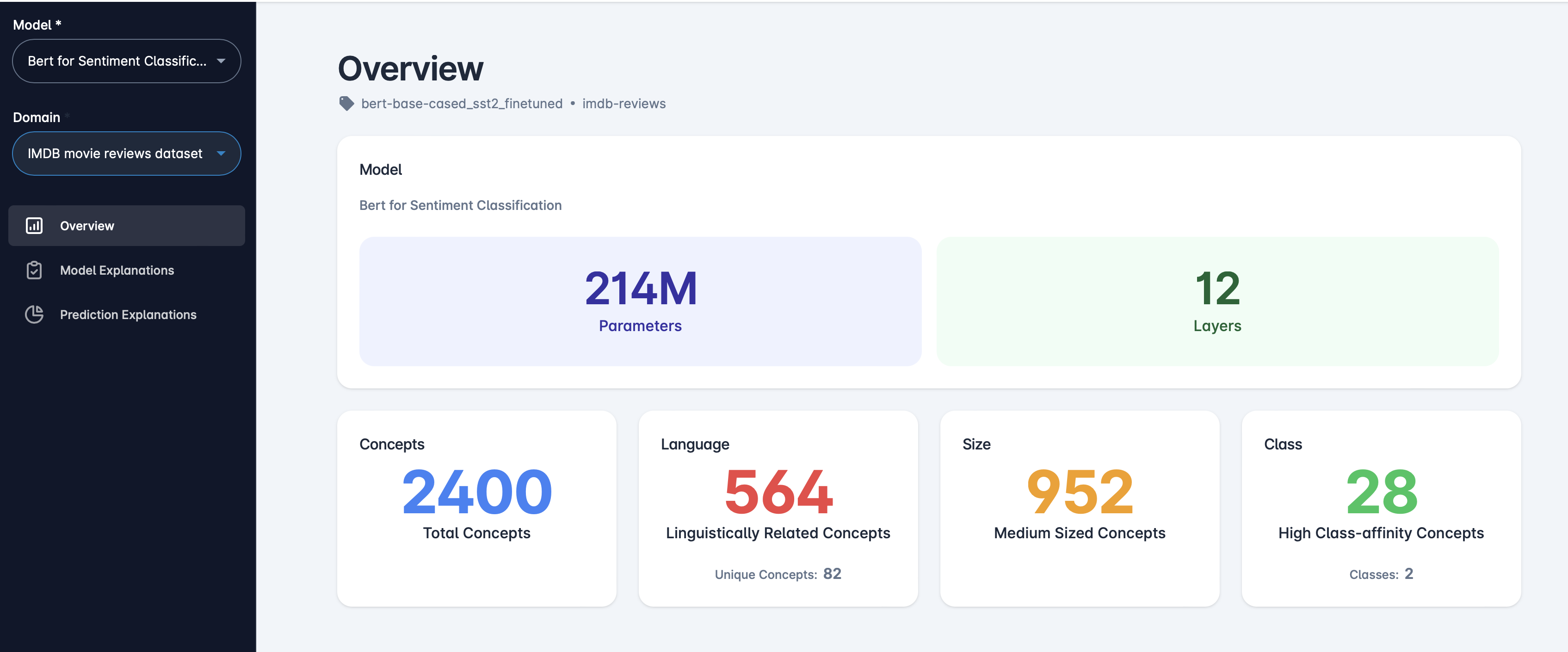}
    \caption{Sample overview page, providing high level statistics at a glance.}
    \label{fig:overview-page}
\end{figure*}

\subsection{Concept Alignment}
The second component uses an alignment framework proposed by \newcite{sajjad-etal-2022-analyzing} to align each of the latent concepts to some pre-existing ontology like part-of-speech, semantic tags, WordNet etc. This enables richer explanations for the latent concepts, and also allows for the application to sort all of the concepts based on criteria relevant to the user. For instance, if the user is only interested in morphological latent concepts, the application can easily filter and sort all of the latent concepts based on this property after the alignment has been performed. 

The alignment of a concept to a specific property (e.g. Noun) is done by checking if most of the words (above a certain threshold) in the concept are labeled with that property. For example, $C_{pos} (JJR) = \{greener, taller, happier, \dots \}$ would be aligned to the property of "comparative adjectives" in the POS tagging task, $C_{sem} (MOY) = \{January, February, \dots, December\}$ defines a concept containing months of the year in the semantic tagging task, and $C_{muslim} (names) = \{Ahmed, Muhammad, Karim, Hamdy, \dots \}$ represents a concept of Muslim names. Explanations based on human-defined concepts are not always applicable or available as these models learn very fine-grained hierarchies of knowledge and concepts that are very task-specific, hence not every latent concept is aligned to some pre-existing tag/ontology.

\subsection{Attribution Analysis}

Our first two components are geared towards understanding what the model has learned, however, it does not necessarily imply that this knowledge is utilized during prediction and provides no insight into how these concepts are being used. To bridge this gap, our third component uses \textbf{Integrated Gradients} (IG) \cite{SundararajanTY17}, which is a powerful axiomatic attribution method for deep neural networks that computes the importance of input features and model components based on their contribution to model's prediction. More concretely, IG is used to extract the salient input features (words) used to make a certain prediction, and these salient features are then mapped to latent concepts to expand on the explanation. For example in Figure \ref{fig:prediction-explanations} highlights ``captures" to be the most salient input feature used in predicting the sentiment of the sentence.



\begin{figure*}[!ht]
    \centering
    \includegraphics[width=0.99\linewidth]{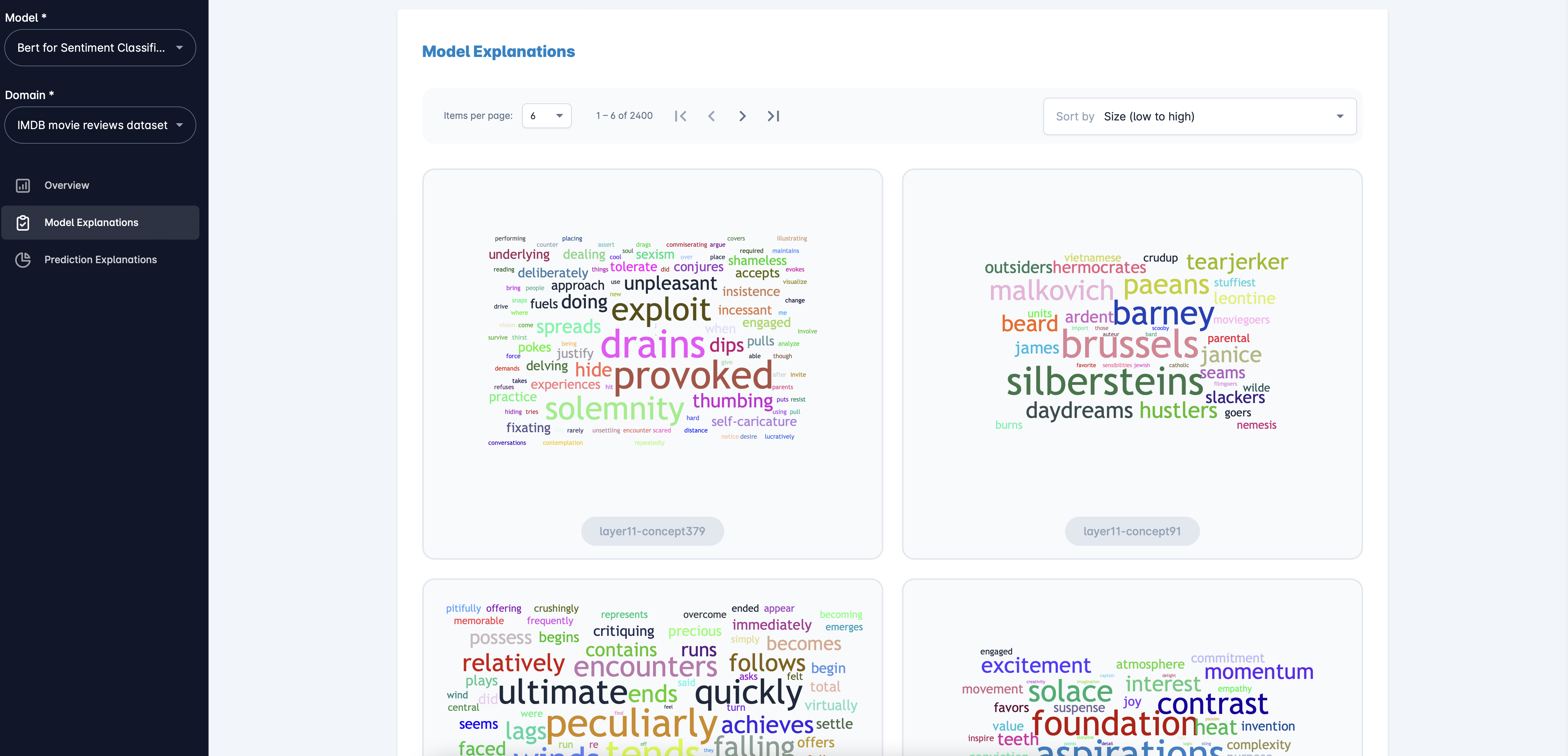}
    \caption{The model-explanation page showing latent concepts for the selected model and domain. Sorting and pagination controls allow a user to effectively browse and analyze concepts learned by the model.}
    \label{fig:model-explanations}
\end{figure*}

\section{Frontend Views}

The goal of \textbf{NxPlain} is to provide an easy method for users to extract and analyze latent knowledge learned within a deep NLP model and connect them to the prediction. The \texttt{Frontend} helps achieve this goal by providing a intuitive yet powerful GUI that can be used to interact with a model's latent concepts and predictions. The user can upload a model and a corpus that they want to analyze. The computational queue of the application discovers latent concepts and aligns them using the components mentioned in Section \ref{sec:research-modules}. The user can then use the \texttt{Frontend}, where they can switch between three major views:

\subparagraph{Overall view:} This view presents a high-level overview of the concepts learned by the model. Specifically, we can see i) the number of concepts learned, ii) statistics on the concepts aligned with the human-fined concepts, iii) a summary of the size distribution of these concepts, iv) and salient concepts in the data and model. Figure \ref{fig:overview-page} shows a sample overview page for a Sentiment analysis model.

\subparagraph{Model Explanations view:} This view presents the latent concepts in a paginated view, along with controls to sort the concepts. Users can sort the concepts i) by size, ii) by their affinity to the linguistic phenomenon (using the alignments computed earlier), iii) by their relation to the various output classes (in classification models) and iv) by their overall relevance. Each concept is accompanied by a unique label to keep track of important concepts. See Figure \ref{fig:model-explanations} for a sample model explanation view.

\begin{figure*}[t]
    \centering
    \includegraphics[width=0.99\linewidth]{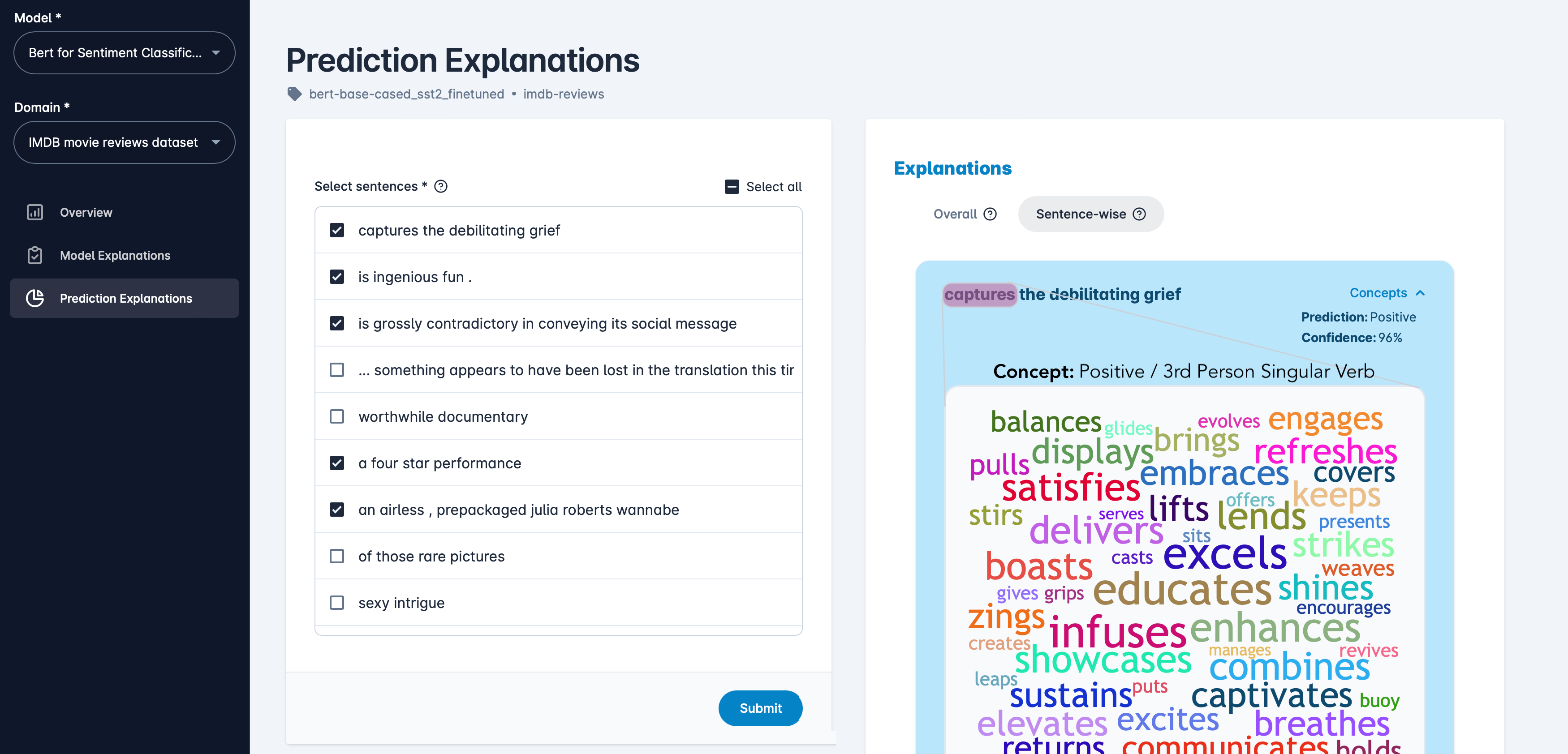}
    \caption{The prediction-explanation page showing latent concepts used during the prediction. The Integrated Gradients method highlights that capture is the most salient word used in the prediction. NxPlain connects it to the concept used along with its label. We observe here that the model used a concept representing positive verbs.}
    \label{fig:prediction-explanations}
\end{figure*}

\subparagraph{Prediction Explanations view:} This view allows the user to look at concepts used in making a prediction and facilitates a deeper view of the behavior of the model on specific sentences. The \textit{attribution analysis} component is used to get a salience map of the input tokens, as well as the matching concepts that contain these tokens in similar contexts. Figure \ref{fig:prediction-explanations} displays the prediction view, where the user can select the sentences that they want to analyze. Here NxPlain shows that ``captures" was the most influential word used by the model to make the prediction. The model used a latent concept representing \textit{positive verbs} to make the prediction. 

\section{Related Work}
\label{sec:related_work}

\subsection{Toolkits}

A number of toolkits have been made available to carry  out analysis of neural network models. Google’s What-If tool \cite{Wexler_2019} inspects machine learning models and provides users an insight into the trained model based on the predictions. Seq2Seq-Vis \cite{strobelt-etal-2018-debugging} enables the user to trace back the prediction decisions to the input in NMT models. Captum \cite{Captum} provides generic implementations of a number of gradient and perturbation-based attribution algorithms. NeuroX \cite{neurox-aaai19:demo} and Ecco~\cite{alammar2021ecco} use probing classifiers to examine the representations pre-trained language models. ConceptX \cite{alam2023conceptX} provides a framework for analyzing and annotating latent concepts in pre-trained language models. \citet{tenney-etal-2020-language} facilitates debugging of pLMs through interactive visualizations. Our work is different from these toolkits. Our toolkit bridges the gap between representation analysis and causation by using attribution-based method. NxPlain provides enriched explanations using traditional linguistic knowledge and human-defined ontologies. 

\subsection{Research Works}
A large number of studies primarily focus on understanding the knowledge learned within a trained model. Researchers have proposed numerous analysis frameworks such as diagnostic classifiers \cite{belinkov:2017:acl, hupkes2018visualisation}, corpus analysis \cite{kadar2016representation,poerner-etal-2018-interpretable,Na-ICLR}, linguistic correlation analysis \cite{dalvi:2019:AAAI,lakretz-etal-2019-emergence}. A plethora of work has been carried out using these analyses frameworks to analyze what concepts are learned within the representations through relevant extrinsic phenomenon varying from word morphology \cite{vylomova2016word, belinkov:2017:acl,dalvi:2017:ijcnlp} to high level concepts such as structure \cite{shi-padhi-knight:2016:EMNLP2016,linzen_tacl,durrani-etal-2019-one}  and semantics \cite{qian-qiu-huang:2016:P16-11, belinkov:2017:ijcnlp} or more generic properties such as sentence length \cite{adi2016fine,bau2018identifying}. 

While the work done on representation analysis unwraps interesting insights about the knowledge learned within the network and how it is preserved, it's only limited to human-defined concepts. More recent work has discovered that these models capture novel ontologies \cite{michael-etal-2020-asking,dalvi2022discovering,Yao-Latent} learning linguistic concepts \cite{sajjad-etal-2022-analyzing}, as well as the task-specific concepts \cite{durrani-EMNLP-22} that emerge as the pre-trained language models are fine-tuned towards a task.

Another line of work in interpretability focuses on attribution analysis that characterizes the role of model components and input features towards a specific prediction~\cite{linzen_tacl,gulordava-etal-2018-colorless,marvin-linzen-2018-targeted}. The explanations are categorized based on two aspects: local or global~\cite{Guidotti}. The former gives a view of explanation at a level of individual instance \cite{LIME,alvarez-melis-jaakkola-2017-causal}, whereas the latter explains the general behavior of the model at corpus level \cite{pryzant-etal-2018-interpretable,prollochs-etal-2019-learning}.

\section{Conclusion}

We presented \textbf{NxPlain}, a web-app for connecting concept analysis with model prediction. The application bridges \textit{representation analysis} and \textit{attribution analysis} to better explain the models' predictions, and provides a intuitive, yet powerful graphical interface to explore the knowledge learned by a model, and also to pinpoint the knowledge used in specific predictions. In the future, we plan to enable human-in-the-loop to enhance concept alignment, as well as incorporate feedback into the explanation system. A hosted version of the application can be accessed at \url{https://nxplain.qcri.org}.

\bibliography{anthology,custom}

\begin{thebibliography}{44}
\expandafter\ifx\csname natexlab\endcsname\relax\def\natexlab#1{#1}\fi

\bibitem[{Adi et~al.(2016)Adi, Kermany, Belinkov, Lavi, and
  Goldberg}]{adi2016fine}
Yossi Adi, Einat Kermany, Yonatan Belinkov, Ofer Lavi, and Yoav Goldberg. 2016.
\newblock {Fine-grained Analysis of Sentence Embeddings Using Auxiliary
  Prediction Tasks}.
\newblock \emph{arXiv preprint arXiv:1608.04207}.

\bibitem[{Alam et~al.(2023)Alam, Dalvi, Durrani, Sajjad, Khan, and
  Xu}]{alam2023conceptX}
Firoj Alam, Fahim Dalvi, Nadir Durrani, Hassan Sajjad, Abdul~Rafae Khan, and
  Jia Xu. 2023.
\newblock Conceptx: A framework for latent concept analysis.
\newblock \emph{Proceedings of the AAAI Conference on Artificial Intelligence}.

\bibitem[{Alammar(2021)}]{alammar2021ecco}
J~Alammar. 2021.
\newblock Ecco: an open source library for the explainability of transformer
  language models.
\newblock In \emph{Proceedings of the 59th Annual Meeting of the Association
  for Computational Linguistics and the 11th International Joint Conference on
  Natural Language Processing: System Demonstrations}, pages 249--257.

\bibitem[{Alvarez-Melis and
  Jaakkola(2017)}]{alvarez-melis-jaakkola-2017-causal}
David Alvarez-Melis and Tommi Jaakkola. 2017.
\newblock \href {https://doi.org/10.18653/v1/D17-1042} {A causal framework for
  explaining the predictions of black-box sequence-to-sequence models}.
\newblock In \emph{Proceedings of the 2017 Conference on Empirical Methods in
  Natural Language Processing}, pages 412--421, Copenhagen, Denmark.
  Association for Computational Linguistics.

\bibitem[{Bau et~al.(2019)Bau, Belinkov, Sajjad, Durrani, Dalvi, and
  Glass}]{bau2018identifying}
Anthony Bau, Yonatan Belinkov, Hassan Sajjad, Nadir Durrani, Fahim Dalvi, and
  James Glass. 2019.
\newblock \href {https://openreview.net/forum?id=H1z-PsR5KX} {Identifying and
  controlling important neurons in neural machine translation}.
\newblock In \emph{Proceedings of the Seventh International Conference on
  Learning Representations}, ICLR~'19, New Orleans, USA.

\bibitem[{Belinkov et~al.(2017{\natexlab{a}})Belinkov, Durrani, Dalvi, Sajjad,
  and Glass}]{belinkov:2017:acl}
Yonatan Belinkov, Nadir Durrani, Fahim Dalvi, Hassan Sajjad, and James Glass.
  2017{\natexlab{a}}.
\newblock \href {https://doi.org/10.18653/v1/P17-1080} {What do neural machine
  translation models learn about morphology?}
\newblock In \emph{Proceedings of the 55th Annual Meeting of the Association
  for Computational Linguistics}, ACL~'17, pages 861--872, Vancouver, Canada.
  Association for Computational Linguistics.

\bibitem[{Belinkov et~al.(2020)Belinkov, Durrani, Dalvi, Sajjad, and
  Glass}]{belinkov-etal-2020-analysis}
Yonatan Belinkov, Nadir Durrani, Fahim Dalvi, Hassan Sajjad, and James Glass.
  2020.
\newblock On the linguistic representational power of neural machine
  translation models.
\newblock \emph{Computational Linguistics}, 45(1):1--57.

\bibitem[{Belinkov et~al.(2017{\natexlab{b}})Belinkov, M{\`a}rquez, Sajjad,
  Durrani, Dalvi, and Glass}]{belinkov:2017:ijcnlp}
Yonatan Belinkov, Llu{\'\i}s M{\`a}rquez, Hassan Sajjad, Nadir Durrani, Fahim
  Dalvi, and James Glass. 2017{\natexlab{b}}.
\newblock \href {https://aclanthology.org/I17-1001} {Evaluating layers of
  representation in neural machine translation on part-of-speech and semantic
  tagging tasks}.
\newblock In \emph{Proceedings of the Eighth International Joint Conference on
  Natural Language Processing}, IJCNLP~'17, pages 1--10, Taipei, Taiwan. Asian
  Federation of Natural Language Processing.

\bibitem[{Dalvi et~al.(2019{\natexlab{a}})Dalvi, Durrani, Sajjad, Belinkov,
  Bau, and Glass}]{dalvi:2019:AAAI}
Fahim Dalvi, Nadir Durrani, Hassan Sajjad, Yonatan Belinkov, D.~Anthony Bau,
  and James Glass. 2019{\natexlab{a}}.
\newblock What is one grain of sand in the desert? analyzing individual neurons
  in deep nlp models.
\newblock In \emph{Proceedings of the Thirty-Third AAAI Conference on
  Artificial Intelligence}, AAAI~'19, pages 6309--6317, Honolulu, Hawaii, USA.
  AAAI.

\bibitem[{Dalvi et~al.(2017)Dalvi, Durrani, Sajjad, Belinkov, and
  Vogel}]{dalvi:2017:ijcnlp}
Fahim Dalvi, Nadir Durrani, Hassan Sajjad, Yonatan Belinkov, and Stephan Vogel.
  2017.
\newblock {Understanding and Improving Morphological Learning in the Neural
  Machine Translation Decoder}.
\newblock In \emph{Proceedings of the 8th International Joint Conference on
  Natural Language Processing (IJCNLP)}.

\bibitem[{Dalvi et~al.(2022)Dalvi, Khan, Alam, Durrani, Xu, and
  Sajjad}]{dalvi2022discovering}
Fahim Dalvi, Abdul~Rafae Khan, Firoj Alam, Nadir Durrani, Jia Xu, and Hassan
  Sajjad. 2022.
\newblock \href {https://openreview.net/forum?id=POTMtpYI1xH} {Discovering
  latent concepts learned in {BERT}}.
\newblock In \emph{Proceedings of the Tenth International Conference on
  Learning Representations}, ICLR~'22, Online.

\bibitem[{Dalvi et~al.(2019{\natexlab{b}})Dalvi, Nortonsmith, Bau, Belinkov,
  Sajjad, Durrani, and Glass}]{neurox-aaai19:demo}
Fahim Dalvi, Avery Nortonsmith, D.~Anthony Bau, Yonatan Belinkov, Hassan
  Sajjad, Nadir Durrani, and James Glass. 2019{\natexlab{b}}.
\newblock Neurox: A toolkit for analyzing individual neurons in neural
  networks.
\newblock In \emph{Proceedings of the AAAI Conference on Artificial
  Intelligence}, AAAI~'19, pages 9851--9852, Honolulu, USA.

\bibitem[{Danilevsky et~al.(2020)Danilevsky, Qian, Aharonov, Katsis, Kawas, and
  Sen}]{danilevsky-etal-2020-survey}
Marina Danilevsky, Kun Qian, Ranit Aharonov, Yannis Katsis, Ban Kawas, and
  Prithviraj Sen. 2020.
\newblock \href {https://www.aclweb.org/anthology/2020.aacl-main.46} {A survey
  of the state of explainable {AI} for natural language processing}.
\newblock In \emph{Proceedings of the 1st Conference of the Asia-Pacific
  Chapter of the Association for Computational Linguistics and the 10th
  International Joint Conference on Natural Language Processing}, pages
  447--459, Suzhou, China. Association for Computational Linguistics.

\bibitem[{Durrani et~al.(2019)Durrani, Dalvi, Sajjad, Belinkov, and
  Nakov}]{durrani-etal-2019-one}
Nadir Durrani, Fahim Dalvi, Hassan Sajjad, Yonatan Belinkov, and Preslav Nakov.
  2019.
\newblock \href {https://doi.org/10.18653/v1/N19-1154} {One size does not fit
  all: Comparing {NMT} representations of different granularities}.
\newblock In \emph{Proceedings of the 2019 Conference of the North {A}merican
  Chapter of the Association for Computational Linguistics: Human Language
  Technologies}, ACL~'19, pages 1504--1516, Minneapolis, Minnesota, USA.
  Association for Computational Linguistics.

\bibitem[{Durrani et~al.(2022)Durrani, Sajjad, Dalvi, and
  Alam}]{durrani-EMNLP-22}
Nadir Durrani, Hassan Sajjad, Fahim Dalvi, and Firoj Alam. 2022.
\newblock On the transformation of latent space in fine-tuned nlp models.
\newblock In \emph{Proceedings of the 2022 Conference on Empirical Methods in
  Natural Language Processing}, EMNLP, Abu Dhabi, UAE. Association for
  Computational Linguistics.

\bibitem[{Elazar et~al.(2021)Elazar, Ravfogel, Jacovi, and
  Goldberg}]{elazar-etal-2021-amnesic}
Yanai Elazar, Shauli Ravfogel, Alon Jacovi, and Yoav Goldberg. 2021.
\newblock \href {https://doi.org/10.1162/tacl_a_00359} {Amnesic probing:
  Behavioral explanation with amnesic counterfactuals}.
\newblock \emph{Transactions of the Association for Computational Linguistics},
  9:160--175.

\bibitem[{Feder et~al.(2021)Feder, Oved, Shalit, and
  Reichart}]{feder-etal-2021-causalm}
Amir Feder, Nadav Oved, Uri Shalit, and Roi Reichart. 2021.
\newblock \href {https://doi.org/10.1162/coli_a_00404} {{C}ausa{LM}: Causal
  model explanation through counterfactual language models}.
\newblock \emph{Computational Linguistics}, 47(2):333--386.

\bibitem[{Fu and Lapata(2022)}]{Yao-Latent}
Yao Fu and Mirella Lapata. 2022.
\newblock \href {https://doi.org/10.48550/ARXIV.2206.01512} {Latent topology
  induction for understanding contextualized representations}.

\bibitem[{Gowda and Krishna(1978)}]{gowda1978agglomerative}
K~Chidananda Gowda and G~Krishna. 1978.
\newblock Agglomerative clustering using the concept of mutual nearest
  neighbourhood.
\newblock \emph{Pattern recognition}, 10(2):105--112.

\bibitem[{Guidotti et~al.(2018)Guidotti, Monreale, Turini, Pedreschi, and
  Giannotti}]{Guidotti}
Riccardo Guidotti, Anna Monreale, Franco Turini, Dino Pedreschi, and Fosca
  Giannotti. 2018.
\newblock \href {http://arxiv.org/abs/1802.01933} {A survey of methods for
  explaining black box models}.
\newblock \emph{CoRR}, abs/1802.01933.

\bibitem[{Gulordava et~al.(2018)Gulordava, Bojanowski, Grave, Linzen, and
  Baroni}]{gulordava-etal-2018-colorless}
Kristina Gulordava, Piotr Bojanowski, Edouard Grave, Tal Linzen, and Marco
  Baroni. 2018.
\newblock \href {https://doi.org/10.18653/v1/N18-1108} {Colorless green
  recurrent networks dream hierarchically}.
\newblock In \emph{Proceedings of the 2018 Conference of the North {A}merican
  Chapter of the Association for Computational Linguistics: Human Language
  Technologies}, NAACL~'18, pages 1195--1205, New Orleans, Louisiana, USA.
  Association for Computational Linguistics.

\bibitem[{Hupkes et~al.(2018)Hupkes, Veldhoen, and
  Zuidema}]{hupkes2018visualisation}
Dieuwke Hupkes, Sara Veldhoen, and Willem Zuidema. 2018.
\newblock Visualisation and 'diagnostic classifiers' reveal how recurrent and
  recursive neural networks process hierarchical structure.
\newblock \emph{arXiv:1711.10203}.

\bibitem[{K{\'a}d{\'a}r et~al.(2017)K{\'a}d{\'a}r, Chrupa{\l}a, and
  Alishahi}]{kadar2016representation}
Akos K{\'a}d{\'a}r, Grzegorz Chrupa{\l}a, and Afra Alishahi. 2017.
\newblock Representation of linguistic form and function in recurrent neural
  networks.
\newblock \emph{Computational Linguistics}, 43(4):761--780.

\bibitem[{Kokhlikyan et~al.(2020)Kokhlikyan, Miglani, Martin, Wang, Alsallakh,
  Reynolds, Melnikov, Kliushkina, Araya, Yan, and Reblitz-Richardson}]{Captum}
Narine Kokhlikyan, Vivek Miglani, Miguel Martin, Edward Wang, Bilal Alsallakh,
  Jonathan Reynolds, Alexander Melnikov, Natalia Kliushkina, Carlos Araya, Siqi
  Yan, and Orion Reblitz-Richardson. 2020.
\newblock \href {https://doi.org/10.48550/ARXIV.2009.07896} {Captum: A unified
  and generic model interpretability library for pytorch}.

\bibitem[{Lakretz et~al.(2019)Lakretz, Kruszewski, Desbordes, Hupkes, Dehaene,
  and Baroni}]{lakretz-etal-2019-emergence}
Yair Lakretz, German Kruszewski, Theo Desbordes, Dieuwke Hupkes, Stanislas
  Dehaene, and Marco Baroni. 2019.
\newblock \href {https://doi.org/10.18653/v1/N19-1002} {The emergence of number
  and syntax units in {LSTM} language models}.
\newblock In \emph{Proceedings of the 2019 Conference of the North {A}merican
  Chapter of the Association for Computational Linguistics: Human Language
  Technologies, Volume 1 (Long and Short Papers)}, pages 11--20, Minneapolis,
  Minnesota. Association for Computational Linguistics.

\bibitem[{Linzen et~al.(2016)Linzen, Dupoux, and Goldberg}]{linzen_tacl}
Tal Linzen, Emmanuel Dupoux, and Yoav Goldberg. 2016.
\newblock {Assessing the ability of LSTMs to learn syntax-sensitive
  dependencies}.
\newblock \emph{Transactions of the Association for Computational Linguistics},
  4:521– 535.

\bibitem[{Marvin and Linzen(2018)}]{marvin-linzen-2018-targeted}
Rebecca Marvin and Tal Linzen. 2018.
\newblock \href {https://doi.org/10.18653/v1/D18-1151} {Targeted syntactic
  evaluation of language models}.
\newblock In \emph{Proceedings of the 2018 Conference on Empirical Methods in
  Natural Language Processing}, pages 1192--1202, Brussels, Belgium.
  Association for Computational Linguistics.

\bibitem[{Michael et~al.(2020)Michael, Botha, and
  Tenney}]{michael-etal-2020-asking}
Julian Michael, Jan~A. Botha, and Ian Tenney. 2020.
\newblock \href {https://doi.org/10.18653/v1/2020.emnlp-main.552} {Asking
  without telling: Exploring latent ontologies in contextual representations}.
\newblock In \emph{Proceedings of the 2020 Conference on Empirical Methods in
  Natural Language Processing}, EMNLP~'20, pages 6792--6812, Online.
  Association for Computational Linguistics.

\bibitem[{Na et~al.(2019)Na, Choe, Lee, and Kim}]{Na-ICLR}
Seil Na, Yo~Joong Choe, Dong{-}Hyun Lee, and Gunhee Kim. 2019.
\newblock \href {http://arxiv.org/abs/1902.07249} {Discovery of natural
  language concepts in individual units of {CNNs}}.
\newblock \emph{CoRR}, abs/1902.07249.

\bibitem[{Poerner et~al.(2018)Poerner, Roth, and
  Sch{\"u}tze}]{poerner-etal-2018-interpretable}
Nina Poerner, Benjamin Roth, and Hinrich Sch{\"u}tze. 2018.
\newblock \href {https://doi.org/10.18653/v1/W18-5437} {Interpretable textual
  neuron representations for {NLP}}.
\newblock In \emph{Proceedings of the 2018 {EMNLP} Workshop {B}lackbox{NLP}:
  Analyzing and Interpreting Neural Networks for {NLP}}, pages 325--327,
  Brussels, Belgium. Association for Computational Linguistics.

\bibitem[{Pr{\"o}llochs et~al.(2019)Pr{\"o}llochs, Feuerriegel, and
  Neumann}]{prollochs-etal-2019-learning}
Nicolas Pr{\"o}llochs, Stefan Feuerriegel, and Dirk Neumann. 2019.
\newblock \href {https://doi.org/10.18653/v1/N19-1038} {Learning interpretable
  negation rules via weak supervision at document level: A reinforcement
  learning approach}.
\newblock In \emph{Proceedings of the 2019 Conference of the North {A}merican
  Chapter of the Association for Computational Linguistics: Human Language
  Technologies, Volume 1 (Long and Short Papers)}, pages 407--413, Minneapolis,
  Minnesota. Association for Computational Linguistics.

\bibitem[{Pryzant et~al.(2018)Pryzant, Basu, and
  Sone}]{pryzant-etal-2018-interpretable}
Reid Pryzant, Sugato Basu, and Kazoo Sone. 2018.
\newblock \href {https://doi.org/10.18653/v1/W18-5415} {Interpretable neural
  architectures for attributing an ad{'}s performance to its writing style}.
\newblock In \emph{Proceedings of the 2018 {EMNLP} Workshop {B}lackbox{NLP}:
  Analyzing and Interpreting Neural Networks for {NLP}}, pages 125--135,
  Brussels, Belgium. Association for Computational Linguistics.

\bibitem[{Qian et~al.(2016)Qian, Qiu, and Huang}]{qian-qiu-huang:2016:P16-11}
Peng Qian, Xipeng Qiu, and Xuanjing Huang. 2016.
\newblock \href {http://www.aclweb.org/anthology/P16-1140} {{Investigating
  Language Universal and Specific Properties in Word Embeddings}}.
\newblock In \emph{Proceedings of the 54th Annual Meeting of the Association
  for Computational Linguistics}, ACL~'16, pages 1478--1488, Berlin, Germany.
  Association for Computational Linguistics.

\bibitem[{Ribeiro et~al.(2016)Ribeiro, Singh, and Guestrin}]{LIME}
Marco~Tulio Ribeiro, Sameer Singh, and Carlos Guestrin. 2016.
\newblock "why should {I} trust you?": Explaining the predictions of any
  classifier.
\newblock In \emph{Proceedings of the 22nd {ACM} {SIGKDD} International
  Conference on Knowledge Discovery and Data Mining, San Francisco, CA, USA,
  August 13-17, 2016}, pages 1135--1144.

\bibitem[{Sajjad et~al.(2021)Sajjad, Durrani, and Dalvi}]{neuronSurvey}
Hassan Sajjad, Nadir Durrani, and Fahim Dalvi. 2021.
\newblock \href {http://arxiv.org/abs/2108.13138} {{Neuron-level Interpretation
  of Deep NLP Models: A Survey}}.
\newblock \emph{CoRR}, abs/2108.13138.

\bibitem[{Sajjad et~al.(2022)Sajjad, Durrani, Dalvi, Alam, Khan, and
  Xu}]{sajjad-etal-2022-analyzing}
Hassan Sajjad, Nadir Durrani, Fahim Dalvi, Firoj Alam, Abdul Khan, and Jia Xu.
  2022.
\newblock \href {https://doi.org/10.18653/v1/2022.naacl-main.225} {Analyzing
  encoded concepts in transformer language models}.
\newblock In \emph{Proceedings of the 2022 Conference of the North American
  Chapter of the Association for Computational Linguistics: Human Language
  Technologies}, pages 3082--3101, Seattle, United States. Association for
  Computational Linguistics.

\bibitem[{Shi et~al.(2016)Shi, Padhi, and
  Knight}]{shi-padhi-knight:2016:EMNLP2016}
Xing Shi, Inkit Padhi, and Kevin Knight. 2016.
\newblock Does string-based neural {MT} learn source syntax?
\newblock In \emph{Proceedings of the 2016 Conference on Empirical Methods in
  Natural Language Processing}, EMNLP~'16, pages 1526--1534, Austin, TX, USA.

\bibitem[{Strobelt et~al.(2018)Strobelt, Gehrmann, Behrisch, Perer, Pfister,
  and Rush}]{strobelt-etal-2018-debugging}
Hendrik Strobelt, Sebastian Gehrmann, Michael Behrisch, Adam Perer, Hanspeter
  Pfister, and Alexander Rush. 2018.
\newblock \href {https://doi.org/10.18653/v1/W18-5451} {Debugging
  sequence-to-sequence models with {S}eq2{S}eq-vis}.
\newblock In \emph{Proceedings of the 2018 {EMNLP} Workshop {B}lackbox{NLP}:
  Analyzing and Interpreting Neural Networks for {NLP}}, pages 368--370,
  Brussels, Belgium. Association for Computational Linguistics.

\bibitem[{Sundararajan et~al.(2017)Sundararajan, Taly, and
  Yan}]{SundararajanTY17}
Mukund Sundararajan, Ankur Taly, and Qiqi Yan. 2017.
\newblock \href {http://arxiv.org/abs/1703.01365} {Axiomatic attribution for
  deep networks}.
\newblock \emph{CoRR}, abs/1703.01365.

\bibitem[{Tenney et~al.(2019)Tenney, Das, and Pavlick}]{tenney-etal-2019-bert}
Ian Tenney, Dipanjan Das, and Ellie Pavlick. 2019.
\newblock \href {https://doi.org/10.18653/v1/P19-1452} {{BERT} rediscovers the
  classical {NLP} pipeline}.
\newblock In \emph{Proceedings of the 57th Annual Meeting of the Association
  for Computational Linguistics}, pages 4593--4601, Florence, Italy.
  Association for Computational Linguistics.

\bibitem[{Tenney et~al.(2020)Tenney, Wexler, Bastings, Bolukbasi, Coenen,
  Gehrmann, Jiang, Pushkarna, Radebaugh, Reif, and
  Yuan}]{tenney-etal-2020-language}
Ian Tenney, James Wexler, Jasmijn Bastings, Tolga Bolukbasi, Andy Coenen,
  Sebastian Gehrmann, Ellen Jiang, Mahima Pushkarna, Carey Radebaugh, Emily
  Reif, and Ann Yuan. 2020.
\newblock \href {https://doi.org/10.18653/v1/2020.emnlp-demos.15} {The language
  interpretability tool: Extensible, interactive visualizations and analysis
  for {NLP} models}.
\newblock In \emph{Proceedings of the 2020 Conference on Empirical Methods in
  Natural Language Processing: System Demonstrations}, pages 107--118, Online.
  Association for Computational Linguistics.

\bibitem[{Vylomova et~al.(2017)Vylomova, Cohn, He, and
  Haffari}]{vylomova2016word}
Ekaterina Vylomova, Trevor Cohn, Xuanli He, and Gholamreza Haffari. 2017.
\newblock \href {https://doi.org/10.18653/v1/W17-4115} {Word representation
  models for morphologically rich languages in neural machine translation}.
\newblock In \emph{Proceedings of the First Workshop on Subword and Character
  Level Models in {NLP}}, pages 103--108, Copenhagen, Denmark. Association for
  Computational Linguistics.

\bibitem[{Wexler et~al.(2019)Wexler, Pushkarna, Bolukbasi, Wattenberg, Viegas,
  and Wilson}]{Wexler_2019}
James Wexler, Mahima Pushkarna, Tolga Bolukbasi, Martin Wattenberg, Fernanda
  Viegas, and Jimbo Wilson. 2019.
\newblock \href {https://doi.org/10.1109/tvcg.2019.2934619} {The what-if tool:
  Interactive probing of machine learning models}.
\newblock \emph{{IEEE} Transactions on Visualization and Computer Graphics},
  pages 1--1.

\bibitem[{Wolf et~al.(2020)Wolf, Debut, Sanh, Chaumond, Delangue, Moi, Cistac,
  Rault, Louf, Funtowicz, Davison, Shleifer, von Platen, Ma, Jernite, Plu, Xu,
  Scao, Gugger, Drame, Lhoest, and Rush}]{wolf-etal-2020-transformers}
Thomas Wolf, Lysandre Debut, Victor Sanh, Julien Chaumond, Clement Delangue,
  Anthony Moi, Pierric Cistac, Tim Rault, Rémi Louf, Morgan Funtowicz, Joe
  Davison, Sam Shleifer, Patrick von Platen, Clara Ma, Yacine Jernite, Julien
  Plu, Canwen Xu, Teven~Le Scao, Sylvain Gugger, Mariama Drame, Quentin Lhoest,
  and Alexander~M. Rush. 2020.
\newblock \href {https://www.aclweb.org/anthology/2020.emnlp-demos.6}
  {Transformers: State-of-the-art natural language processing}.
\newblock In \emph{Proceedings of the 2020 Conference on Empirical Methods in
  Natural Language Processing: System Demonstrations}, pages 38--45, Online.
  Association for Computational Linguistics.

\end{thebibliography}
\bibliographystyle{acl_natbib}

\appendix



\end{document}